\documentclass[letterpaper]{article} 
\usepackage[preprint]{aaai2027}  
\usepackage[hyphens]{url}  
\usepackage{graphicx} 
\usepackage{natbib}  
\usepackage{caption} 
\usepackage{algorithm}
\usepackage{algorithmic}
\usepackage{amssymb}
\usepackage{comment}

\usepackage{newfloat}
\usepackage{listings}
\DeclareCaptionStyle{ruled}{labelfont=normalfont,labelsep=colon,strut=off} 
\floatstyle{ruled}
\newfloat{listing}{tb}{lst}{}
\floatname{listing}{Listing}

\usepackage{booktabs}

\title{Accuracy Does Not Guarantee Human-Likeness: Cross-Domain Human-Centered Benchmark in Monocular Depth Estimation}
\author {
    Yuki Kubota\textsuperscript{\rm 1}\corresponding,
    Taiki Fukiage\textsuperscript{\rm 1},
}
\affiliations {
    Communication Science Laboratories, NTT Inc., \\
    \textsuperscript{\rm 1}3-1 Morinosato-Wakamiya, Atsugi-shi, Kanagawa, 243-0198, Japan \\
    yuki.kubota95@gmail.com
}

\begin{document}

\maketitle

\begin{abstract}
Deep neural networks (DNNs) are increasingly used as functional models of human vision, yet standard monocular depth estimation (MDE) benchmarks largely evaluate physical accuracy rather than behavioral alignment with humans. We introduce a cross-domain behavioral benchmark that aligns newly collected human absolute-distance judgments, physical ground truth, and model predictions at 16 locations in each of 652 KITTI and 654 NYU Depth V2 scenes. Human--model similarity was measured image-wise by partial correlation between scale-and-shift-aligned estimates while controlling for physical depth, and reference results were obtained for 69 diverse MDE models. Human error patterns were highly reliable, and most models showed positive but substantially lower similarity to humans. Human similarity, however, did not increase monotonically with metric accuracy: within each dataset, high-accuracy models converged toward similar residual error structures that remained distinct from humans, while both human-similarity scores and error-space organization were only weakly preserved across datasets. Thus, physical accuracy and human-like behavior are complementary, domain-sensitive dimensions of MDE performance. Our benchmark will provide a reusable basis for assessing future models along both dimensions.
\end{abstract}

\section{Introduction}
\begin{figure*}[htb]
  \centering
  \includegraphics[keepaspectratio,width=0.95\linewidth]{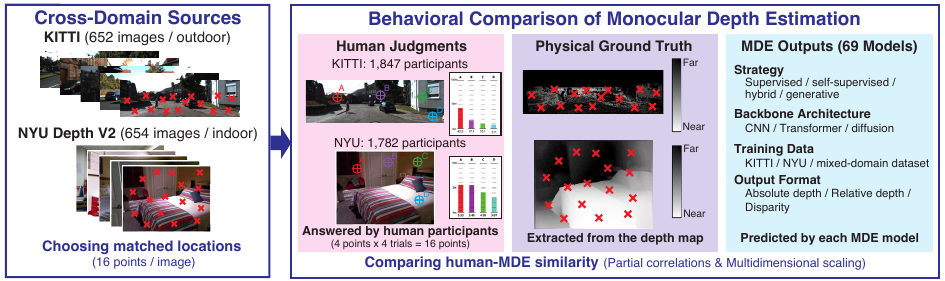}
  \caption{Overview of the proposed cross-domain behavioral benchmark. Human distance judgments, physical ground-truth depth, and predictions from 69 reference MDE models are aligned at the same 16 locations in KITTI and NYU Depth V2. Error-pattern similarity was quantified using Pearson partial correlations and multidimensional scaling.}
  \label{fig:key-figure}
\end{figure*}

Deep neural networks (DNNs) have become important not only as high-performing task solvers, but also as candidate computational models of human perception and cognition. Comparing DNN behavior with human judgments allows researchers to examine whether task optimization gives rise to representations and response patterns that approximate those of the human perceptual system~\cite{wichmann2023deep,zhang2025visualactbench}. Such comparisons serve two complementary purposes. First, they test the adequacy of DNNs as functional models of human visual perception. Second, they reveal behaviorally meaningful differences among models that may be obscured by conventional accuracy-based benchmarks. Human-aligned representations and behaviors have also been associated, in some settings, with greater robustness to distortions and adversarial perturbations~\cite{geirhos2018generalisation,engstrom2019adversarial,wichmann2023deep,sucholutsky2023getting}. Together, these considerations motivate examining whether gains in task performance accompany greater human-likeness, with implications for both cognitive science and artificial intelligence.

The relationship between human and DNN perception has been studied most extensively in object recognition. Early work showed that DNNs optimized for object classification can account for neural and behavioral responses in the primate visual system~\cite{yamins2014performance}. Subsequent studies, however, have revealed a more nuanced picture: DNNs often diverge systematically from human perception, and gains in benchmark accuracy do not necessarily translate into greater alignment with human perceptual judgments~\cite{geirhos2018imagenet,kumar2022better,dehghani2023scaling,hernandez2025vision}.

This alignment-focused research has recently expanded beyond object recognition to mid-level visual functions, including material, depth, 3D shape, and motion estimation~\cite{prokott2021gloss,bae2023study,bonnen2024evaluating,kubota2025human,yang2025huperflow}. Among these domains, spatial perception warrants particular attention as AI systems are increasingly expected to reason about and act in 3D environments~\cite{rajapaksha2024deep,zhou2024comprehensive}. Monocular depth estimation (MDE)  offers a tractable test because human and model errors can both be evaluated against physical depth. Earlier human-annotated datasets instead used human judgments primarily as supervision for model training, without pairing them with physical depth~\cite{chen2016single,lee2022instance}. Recent work enabled such a comparison for indoor scenes and analyzed several components of systematic depth bias~\cite{kubota2025human}. However, the cross-domain consistency of human-like error patterns and the broader error structure relating humans and diverse models remained unclear. 

To address these gaps, we introduce a cross-domain behavioral benchmark for MDE, combining new large-scale human depth judgments at 16 locations per image for both KITTI~\cite{uhrig2017sparsity} and NYU Depth V2~\cite{silberman2012indoor} with reference evaluations of 69 MDE models across outdoor and indoor scene domains. Using this benchmark, joint analyses of human--model and model--model similarity showed that human similarity did not increase monotonically with metric accuracy: within each dataset, the most human-like models occurred at intermediate accuracy levels, whereas high-accuracy models clustered together while remaining distinct from humans; however, this organization was only weakly preserved across datasets. These findings highlight human similarity as a complementary, domain-sensitive evaluation axis beyond metric accuracy, while the benchmark provides a reusable basis for evaluating future MDE models.

\section{Related Work}
\noindent \paragraph{Evaluation of monocular depth estimation models.}
Monocular depth estimation (MDE), which infers 3D structure from a single RGB image, has progressed rapidly with advances in deep learning, from convolutional, Transformer-based, and generative models~\cite{eigen2014depth,li2022depthformer,ranftl2021vision,ke2023repurposing,lavreniuk2023evp}, to large-scale mixed-dataset training and self-supervised learning~\cite{birkl2023midas,oquab2023dinov2,sun2023sc}. Evaluation has largely centered on physical accuracy, using benchmarks such as KITTI~\cite{uhrig2017sparsity} and NYU Depth V2~\cite{silberman2012indoor} and metrics such as AbsRel and RMSE, often after per-image scale alignment to address the inherent scale ambiguity of monocular prediction~\cite{eigen2014depth,ranftl2020towards}. More recent protocols also assess zero-shot cross-dataset generalization~\cite{yang2024depth}. Although these evaluations measure physical accuracy and robustness, they do not test whether current MDE models share the systematic error patterns characteristic of human depth perception. We address this question by comparing human and DNN depth-estimation errors under matched conditions, using physically measured depth as a common reference.

\noindent \paragraph{Human-annotated depth datasets.}
A key strategy for aligning vision models with human perception has been to incorporate human judgments into training and evaluation. In depth estimation, this has often taken the form of ordinal or relative depth annotations, motivated by the fact that humans can reliably judge spatial relationships such as which point or object is closer, even when absolute metric estimates are difficult. Early work showed that models can be trained from pairwise ordinal judgments~\cite{zoran2015learning}, and the Depth in the Wild dataset introduced large-scale human annotations of relative depth for diverse natural images~\cite{chen2016single}. Subsequent studies have further used human-annotated ordinal and occlusion cues to guide depth learning~\cite{lee2022instance}. More broadly, recent benchmarks have begun to evaluate whether vision models exhibit human-like 3D judgments, as in MOCHI for 3D shape inference~\cite{bonnen2024evaluating}. However, most of these approaches do not pair human perceptual estimates with physical metric ground truth, making it difficult to compare residual error patterns between humans and models in depth perception. Recent work addressed this limitation by collecting absolute depth judgments for NYU and examining whether individual MDE models reproduce systematic human biases~\cite{kubota2025human}. The present study shifts the focus to how human--model similarity varies across scene domains and how humans are situated within the broader error structure formed by diverse MDE models. We therefore collected new metric human-depth judgments for both KITTI and NYU Depth V2 and jointly analyzed human--model and model--model similarity using matched cross-dataset comparisons, partial correlations, and multidimensional scaling.

\noindent \paragraph{Human depth perception.}
Research on human depth perception has shifted from treating the visual system as a veridical estimator of physical depth~\cite{landy1995measurement} to emphasizing its systematic distortions from the physical depth structure~\cite{linton2023new}. A widely observed characteristic is depth compression, whereby farther objects are perceived as closer than their physical values~\cite{wagner1985metric,hecht1999compression,linton2023new}. Pictorial depth judgments are also influenced by spatial cues such as vertical position of stimuli and the inferred orientation of surfaces~\cite{vishwanath2005pictures,koenderink2001ambiguity,kubota2022motion}. Such systematic tendencies may reflect learned priors that help resolve the inherent ambiguity of monocular images~\cite{Yang2003-oy}. Together, these findings indicate that human depth-estimation errors are structured rather than random, motivating our investigation of whether contemporary MDE models exhibit similar error patterns.

\section{Methods}
\subsection{Human-annotated depth datasets}
To compare human and model depth estimation under matched conditions, we newly collected human depth judgments for 652 KITTI test scenes~\cite{uhrig2017sparsity} and 654 NYU Depth V2 test scenes~\cite{silberman2012indoor} (``Cross-Domain Sources'' panel in Figure~\ref{fig:key-figure}). Both datasets are widely used monocular depth-estimation benchmarks containing paired RGB images and dense depth maps: KITTI comprises outdoor driving scenes captured with synchronized vehicle-mounted cameras and LiDAR, whereas NYU contains indoor scenes recorded with an RGB-D sensor.

Previous studies have often measured human depth perception using pairwise ordinal judgments or other relative-depth reconstruction tasks~\cite{chen2016single,lee2022instance,wagemans2011measuring}, which do not yield metric estimates directly comparable with physical ground truth. We therefore asked participants to report the \textit{absolute distance} from the camera to evaluation points in each image (``Human Judgments'' panel in Figure~\ref{fig:key-figure}). To improve the reliability of these absolute estimates, participants judged four points simultaneously within each image, providing a local relational context in which to compare the four points and respond their estimates within each scene. Prior work has shown that such absolute judgments are internally reliable and systematically related to ordinal and relative depth judgments~\cite{kubota2025human}.

Because humans cannot feasibly provide pixel-wise depth maps, we sampled 16 evaluation points per image to balance broad scene coverage with sufficient within-image sampling. Points were selected at random while maintaining minimum distances from image borders and segmentation boundaries, then divided into four groups of four for simultaneous judgments within the same image (Appendix~A.1).

We collected depth judgments through crowdsourcing from 1,847 (KITTI) and 1,782 (NYU)  participants aged 20--49 years. Each participant completed an approximately 20-minute experiment covering 28 or 32 trials (7 or 8 images; 112 or 128 points). All points from the same image were assigned to a common participant group to maintain within-image consistency. We excluded participants whose estimates showed low Pearson correlation with the group median. After screening, 1,547 KITTI participants and 1,483 NYU participants remained. The number of valid responses per image averaged 18.87 for KITTI and 18.09 for NYU, ranging across images from 14 to 27 and from 12 to 26, respectively (see Appendix~A.2).

\subsection{Target DNNs: 69 monocular depth estimators}
To systematically compare MDE models with human estimates, we assembled 69 monocular depth estimators spanning diverse training strategies, datasets, backbone architectures, and output forms (``MDE Outputs'' panel in Figure~\ref{fig:key-figure}; see Appendix~A.3 for the full list). Note that we focus exclusively on models trained to minimize the discrepancy with ground-truth physical depth, either directly through ground-truth depth supervision or indirectly through geometric constraints. This selection allowed us to investigate whether human-like error patterns emerge without explicit training on human judgments.

The model set covered supervised, self-supervised, hybrid, and generative approaches. Supervised models learned from paired RGB images and physical depth maps, whereas self-supervised models used monocular videos or stereo pairs with photometric and camera-pose constraints without explicit depth maps. At inference time, both strategies predict pixel-wise depth from a single RGB image. Hybrid models incorporated disparity-based supervision (``hybrid (disparity)'')~\cite{sun2023sc} or semantic representations from large-scale pretraining (``hybrid (semantics)'')~\cite{oquab2023dinov2}; we also included diffusion-based generative models~\cite{lavreniuk2023evp}. Architectures included CNNs, Transformers, CNN--Transformer hybrids, and diffusion-based estimators.

Regarding training data variations, we employed models trained on the KITTI and NYU benchmarks, as well as models trained on other datasets, including indoor datasets (Bonn~\cite{palazzolo2019refusion} and TUM~\cite{sturm2012benchmark}), an outdoor dataset (DDAD~\cite{guizilini20203d}), and several large mixed-domain datasets. Moreover, some models (e.g., MiDaS~\cite{ranftl2021vision} and DINOv2~\cite{oquab2023dinov2}) were pretrained on heterogeneous datasets and, in some cases, subsequently fine-tuned on domain-specific datasets such as KITTI or NYU. We categorize these regimes as ``KITTI + multiple datasets'' and ``NYU + multiple datasets'' in the analysis below.

\subsection{Similarity analysis based on estimation errors}
Because our goal was to compare all 69 MDE models within a common framework despite their different output forms, we adopted scale-and-shift invariant measures for both metric accuracy and error-pattern similarity. For each image, human and model outputs were linearly transformed into depth estimates aligned to physical ground-truth depth at the 16 evaluation points, yielding metric-aligned per-image ``pseudo-absolute'' estimates. For the human analyses, scale recovery was performed separately for each random half-split or bootstrap resample. Human and model accuracy was quantified by RMSE after this per-image alignment, equivalent to scale-shift-invariant RMSE.

Similarity was quantified separately for each image as the Pearson partial correlation between two sets of estimates across the 16 evaluation points, controlling for physical ground-truth depth. This image-wise measure captures correspondence between location-wise residual patterns not linearly explained by ground truth, without conflating them with between-scene differences in scale or depth range. The image-wise correlations were then Fisher $z$-transformed, averaged across images, and back-transformed to obtain an averaged similarity score. (i) Human--human similarity was estimated from 1,000 random observer half-splits. In each split, responses were averaged within each half, image-wise partial correlations were computed between two halves, and the averaged correlation was corrected using the Spearman--Brown formula. The distribution across splits provided confidence intervals of the score. (ii) Human--model similarity was computed between each fixed model output and the mean human estimates; confidence intervals were obtained from 1,000 bootstrap resamples of human responses with replacement, and repeating the scale-recovery and similarity calculation. (iii) Model--model similarity was computed analogously for every model pair and used to construct the similarity matrix for multidimensional scaling.

\subsection{Multidimensional scaling analysis}


To characterize the overall similarity structure among humans and models, we applied metric multidimensional scaling (MDS) to KITTI and NYU. For each dataset, we constructed a similarity matrix comprising the mean human estimates and the 69 MDE models, with each entry defined as the Fisher-$z$-averaged partial correlation between two depth-estimation patterns after controlling for physical ground-truth depth. Pairwise dissimilarities were then defined as $\delta_{ij} = \sqrt{2(1-r_{ij})}$, where $r_{ij}$ denotes the averaged partial correlations, indicating human--model or model--model similarities. We then fitted a two-dimensional metric MDS using eight random initializations and retained the solution with the lowest stress. In the resulting maps, nearby points indicate similar residual error patterns, whereas distant points indicate dissimilar patterns, thereby summarizing human--model and model--model relationships within a common error-pattern space.

\section{Results}
\begin{figure*}[htb]
  \centering
  \includegraphics[keepaspectratio,width=0.95\linewidth]{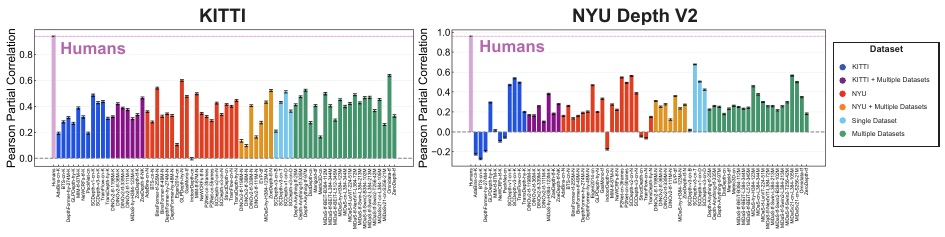}
  \caption{Pearson partial correlations quantifying the similarity of depth-estimation error patterns between humans and MDE models for scale-recovered depth estimates in the KITTI (left) and NYU (right) datasets.}
  \label{fig:similarity-bars}
\end{figure*}

For visualization, MDE models were color-coded into six categories by training data: KITTI only (blue), KITTI fine-tuned after multi-dataset pretraining (purple), NYU only (red), NYU fine-tuned after multi-dataset pretraining (orange), other single datasets (cyan), and other mixed datasets (green). Marker shapes indicate training strategies.

\subsection{Human reliability and reference model scores}
We first established the reliability of the human benchmark and obtained reference human-similarity scores for the 69 MDE models (Figure~\ref{fig:similarity-bars}). The human--human similarity reached  $0.940$ for KITTI and $0.960$ for NYU (half-split correlations prior to Spearman--Brown correction were $0.887$ and $0.923$, respectively), indicating that human depth-estimation errors reflected highly systematic perceptual biases. We then compared each model with this human benchmark. In both the KITTI and NYU datasets, most models showed positive partial correlations with human error patterns, suggesting that models can acquire partially human-like biases even when optimized solely for physical depth accuracy. However, these correlations remained consistently below the human-human similarity level, revealing a clear gap between human perceptual biases and model error patterns.

\subsection{Accuracy does not monotonically predict human similarity}
\begin{figure*}[htb]
  \centering
  \includegraphics[keepaspectratio,width=0.95\linewidth]{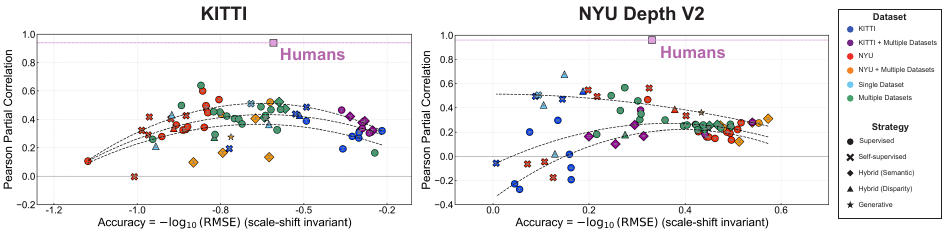}
  \caption{Relationships between metric accuracy and similarity to human error patterns across MDE models in the KITTI (left) and NYU (right) datasets. The horizontal axis represents scale-shift-invariant metric accuracy, and the vertical axis represents similarity to human error patterns. The three curves show the 20\%, 50\%, and 80\% quadratic quantile regression fits.}
  \label{fig:similarity-scatters}
\end{figure*}

We next examined whether greater metric accuracy was associated with greater human similarity (Figure~\ref{fig:similarity-scatters}). Accuracy was defined as the negative logarithm of scale-shift-invariant RMSE, such that higher values indicate greater accuracy while allowing comparison across models with different output scales. Human similarity was quantified as the Pearson partial correlation between human and model depth-estimation error patterns after controlling for physical ground-truth depth. To characterize changes in the conditional distribution of similarity across accuracy levels, we fitted quadratic quantile regressions at the 20\%, 50\%, and 80\% quantiles.

Across both KITTI and NYU, the most accurate models were not the most human-like. Neither the raw observations nor the fitted median curves showed a monotonic increase in human similarity toward the high-accuracy end. The fitted 20th--80th interquantile spread was narrower at the highest observed accuracy than at its maximum, although the reduction was modest for KITTI and pronounced for NYU. It decreased from $0.15$ at a log-accuracy of $-0.60$ to $0.10$ for KITTI, and from $0.84$ at $0.00$ to $0.14$ for NYU. Because similarity was computed after scale-and-shift alignment and after controlling for physical depth, the lower similarity of high-accuracy models indicates that their remaining errors were structured differently from human errors. Overall, these results suggest that, within each dataset, high-accuracy models converged toward a similar level of human similarity that remained below the highest levels observed at intermediate accuracy.


\subsection{Discrepancies in multidimensional scaling maps}
\begin{figure*}[htb]
  \centering
  \includegraphics[keepaspectratio,width=0.95\linewidth]{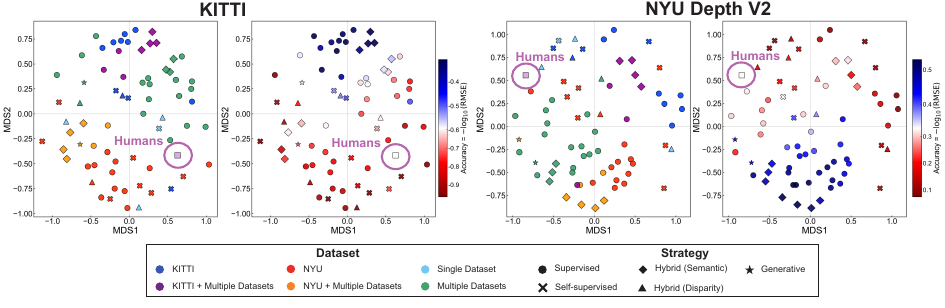}
  \caption{Multidimensional scaling maps of human and MDE model error patterns for KITTI (left) and NYU (right). For each dataset, the left map colors training datasets, whereas the right map colors models according to metric accuracy, with the human accuracy level indicated in white. Marker shapes denote training strategies.}
  \label{fig:similarity-mds}
\end{figure*}

To examine whether high-accuracy models also shared similar broader error structures, we applied multidimensional scaling (MDS) to the pairwise similarity matrix comprising humans and all 69 DNNs (Figure~\ref{fig:similarity-mds}, see Figure~B1 for the full similarity matrices).

In the MDS maps colored by training dataset (the left-hand maps for KITTI and NYU in Figure~\ref{fig:similarity-mds}), models showed partial clustering by training dataset, indicating that training data influenced their residual error structure. However, humans remained separated even from models trained or fine-tuned on the corresponding benchmark, indicating that the dataset-related error structures shared by these models remained distinct from human error structures.

In the accuracy-colored MDS maps (the right-hand maps for KITTI and NYU in Figure~\ref{fig:similarity-mds}), high-accuracy models occupied a common region despite differences in architecture and training configuration. Humans remained separated from this region, suggesting that high-accuracy models shared within-domain residual error structures that were distinct from those of humans. Together with the scatter-plot analysis, these results indicate that physical depth accuracy and human-like perceptual biases capture distinct aspects of model behavior.

\subsection{Cross-dataset consistency of depth-error similarity}
\begin{figure}[htb]
  \centering
  \includegraphics[keepaspectratio,width=1.00\linewidth]{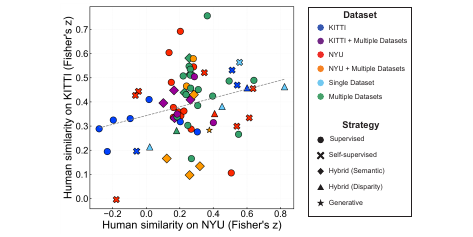}
  \caption{Cross-dataset human--model similarity using Fisher $z$-transformed scores, with NYU on the horizontal axis and KITTI on the vertical axis. The line indicates the ordinary least-squares fit across models.}
  \label{fig:similarity-cross}
\end{figure}

To assess the extent to which error-pattern similarity was preserved across dataset domains, we compared Fisher $z$-transformed human--model similarity scores between NYU and KITTI. Human--model similarity showed a weak positive correlation across the two datasets ($r=0.28$, $p=0.021$) (Figure~\ref{fig:similarity-cross}). The corresponding entries of the model--model similarity matrices were likely weakly correlated across KITTI and NYU (Figure~B1; $r=0.29$, $p<0.0001$). Thus, both a model's similarity to humans and the broader organization of model error patterns were only partially preserved across domains. 

We next tested whether the within-dataset accuracy--similarity relationship generalized across domains by repeating the quadratic quantile-regression analysis using accuracy from one dataset and human similarity from the other (Figure~B2). Unlike the matched-domain analyses, the cross-domain fits showed no consistent change in either the median or the 20th--80th percentile range of human similarity toward the high-accuracy end. Together, these results suggest that human-likeness is not a unitary, domain-invariant property of a model. Rather, it appears to be relational, depending on how model-specific biases interact with the cue and scene statistics of each domain.


\subsection{Human--model similarity across model factors}
We also examined whether human--model similarity varied with training dataset, backbone architecture, and training strategy (see Appendix~B.3 for detailed comparisons). Within matched model families, NYU-trained variants often exhibited higher human similarity than their KITTI-trained counterparts across both test datasets, although this tendency was not universal and was reversed in several model families. Comparisons among CNN, Transformer, and CNN--Transformer hybrid backbones showed no consistent ordering across model groups or test datasets. Some self-supervised models also showed relatively high human similarity, but their scores varied substantially across training-dataset variants. No specific training strategy, including generative or self-supervised learning, consistently produced more human-like error patterns. Overall, these results suggest that human--model similarity reflects the combined influence of training data, architecture, and model family rather than any single design factor.

\subsection{Qualitative results in outdoor and indoor scenes}
\begin{figure*}[htb]
  \centering
  \includegraphics[keepaspectratio,width=0.95\linewidth]{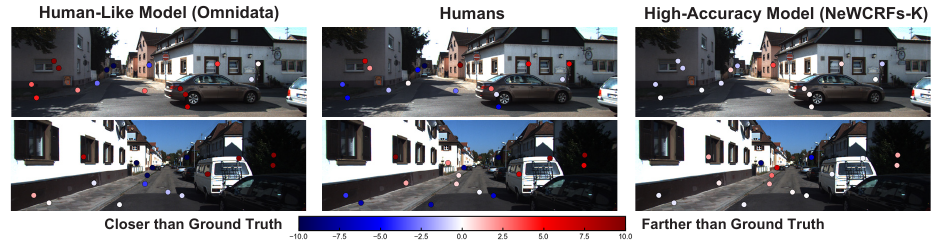}
  \caption{Representative examples of human and MDE model depth-estimation biases for the KITTI dataset.}
  \label{fig:examples-kitti}
\end{figure*}
\begin{figure*}[htb]
  \centering
  \includegraphics[keepaspectratio,width=0.95\linewidth]{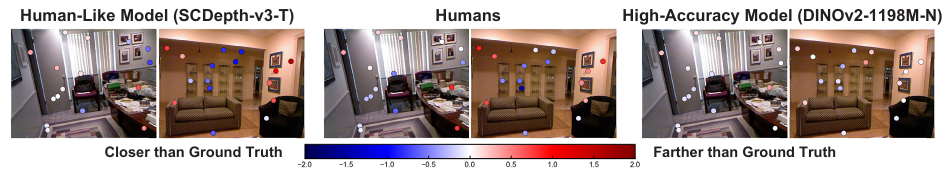}
  \caption{Representative examples of human and MDE model depth-estimation biases for the NYU dataset.}
  \label{fig:examples-nyu}
\end{figure*}

To qualitatively compare human and model depth-estimation error patterns, we overlaid their estimates on KITTI and NYU images. Ten images per dataset were selected by stratified sampling across image-level human--model similarity averaged over all 69 models, rather than according to either displayed comparator (Appendix~B.4). Two examples per dataset are shown here and the remaining eight in Appendix.

For scale-and-shift-aligned visualizations (Figures~\ref{fig:examples-kitti} and B4 for KITTI; Figures~\ref{fig:examples-nyu} and B5 for NYU), humans were compared with one of the human-like models and the highest-accuracy model in each dataset. Marker colors indicate the difference between the aligned estimate and physical ground truth. By aligning each image with a linear scale and shift, these visualizations remove global scale and offset differences and emphasize spatially structured deviations in relative depth that cannot be explained by these linear components, providing an intuitive view of the within-image error organization assessed by our similarity analysis.

Across both datasets, the human-like models reproduced aspects of the sign and spatial arrangement of human residuals that were less apparent in the highest-accuracy models, including corresponding positive residuals on some vertical structures and upright objects. However, the human tendency to assign relatively smaller depths to lower image regions was less consistently reproduced by the MDE models~\cite{vishwanath2005pictures,kubota2022motion}. 

The unaligned metric-output visualizations revealed a complementary bias in absolute depth, most notably depth compression in distant regions (Figures~B6 and B7). Humans and models estimated points near the vanishing region or far end of a scene as closer than the physical ground truth, consistent with the well-documented compression of perceived depth in human vision~\cite{wagner1985metric,wagner2016variations}. Because scale-and-shift alignment and partial correlation remove only linear components, this nonlinear compression can persist in the aligned residuals and may contribute to the positive human--model similarity scores. The remaining spatial differences nevertheless indicate that MDE models reproduce only part of the human bias structure.

\section{Discussion}
We introduced a cross-domain behavioral benchmark for MDE and applied it to 69 models. Our reference evaluation revealed a non-monotonic relationship between physical accuracy and human similarity in both matched-domain scatter plots and MDS maps, paralleling observations in object recognition, where higher ImageNet accuracy can coincide with reduced alignment with human perceptual judgments~\cite{kumar2022better,dehghani2023scaling,hernandez2025vision}.

Interestingly, within each dataset, the quadratic quantile regression and MDS analyses suggest that increasing accuracy is associated with reduced diversity in residual error patterns across models. Despite differences in architecture, training data, and learning strategy, high-accuracy models tended to occupy a common region of the error-pattern space while remaining separated from humans. However, this organization was only weakly preserved across datasets, and the cross-domain accuracy--similarity analyses did not reproduce the matched-domain narrowing at the high-accuracy end. Thus, the observed convergence is better understood as domain-specific organization among high-performing models than as movement toward a single domain-general solution.

The separation between high-accuracy models and humans may arise from how depth cues are learned and constrained. Although the evaluated models differed in architecture and supervision, their objectives ultimately favored physically or geometrically consistent depth. Human cue weighting, by contrast, is calibrated through perceptual experience and interaction with the environment rather than direct optimization against metric depth. Notably, some self-supervised models trained without direct metric-depth supervision showed relatively high human similarity despite lower accuracy than the highest-performing models (Figures~\ref{fig:similarity-scatters} and B3). These observations suggest that the strong optimization pressure to minimize metric-depth error in high-performing models may drive heterogeneous models toward similar residual error structures within a benchmark, but not a fixed, domain-general form of human-likeness.


Our study underscores that progress in depth estimation cannot be fully captured by improvements in metric accuracy alone. Echoing recent calls for richer, multidimensional evaluation frameworks~\cite{wichmann2023deep}, evaluating how models align---or diverge---from human perceptual organization provides an additional lens for characterizing model behavior and assessing their adequacy as functional models of human depth perception. At the same time, the limited cross-domain consistency cautions against treating a human-similarity score from one dataset as a universal model ranking. Evaluations across multiple scene domains can distinguish stable human-like tendencies from domain-specific overlap. The benchmark introduced here provides a reusable basis for such analyses and for identifying computational approaches that are informative for understanding human spatial vision.

\noindent \textbf{Limitations and future work:} Our study is limited by the scope of the datasets and the sparse sampling of evaluation points, motivating future work with denser behavioral measurements and datasets spanning broader variations in scene geometry and depth distribution. More fundamentally, similarity in error patterns provides only one criterion for evaluating whether a DNN constitutes a functional model of human depth perception, as models may reproduce human-like errors while relying on different visual cues or representations. Future studies should therefore evaluate models across statistically diverse datasets and controlled cue manipulations to determine whether they jointly reproduce human accuracy, systematic biases, and context-dependent behavior. It also remains unclear whether models that better reproduce human behavioral profiles across domains exhibit greater out-of-distribution robustness or other generalization benefits.



\section{Conclusions}
This study examined the relationship between matched metric accuracy and human-like behavior across 69 monocular depth estimators evaluated on KITTI and NYU Depth V2. Most MDE models showed positive correlations with human depth-estimation errors, indicating that they reproduced some components of human perceptual bias. However, within each dataset, the most human-like models occurred at intermediate accuracy levels, whereas high-accuracy models converged toward similar residual error structures that remained separated from humans. At the same time, both human--model similarity and the broader model--model similarity structure showed only weak correspondence between KITTI and NYU. Thus, physical depth accuracy and human-like perceptual biases capture distinct aspects of model behavior, while human-likeness itself is relational and domain-sensitive rather than a fixed model attribute. Human-centered benchmarks across multiple domains therefore provide a complementary evaluation axis for identifying perceptually meaningful differences among depth-estimation models that are not revealed by accuracy alone.

\section*{Ethical Statement}
The study protocol was approved by the authors' institutional review board, and the experiment was conducted in accordance with the Declaration of Helsinki. 

\bibliography{ref}

\onecolumn
\appendix
\setcounter{secnumdepth}{2}

\counterwithin{figure}{section}
\renewcommand{\thefigure}{\thesection\arabic{figure}}

\counterwithin{table}{section}
\renewcommand{\thetable}{\thesection\arabic{table}}

\section{Supplemental Methods}
An anonymized version of the human response data and the analysis code used in this study is included in the Code and Data Supplement. These materials will also be released publicly upon publication.

\subsection{Per-image evaluation point sampling}
\label{app:point-sampling}
For each of the 652 evaluation images (KITTI) and 654 evaluation images (NYU), we selected four groups of four evaluation points (sixteen points per image) following three spatial constraints designed to ensure the reliability of the evaluation points. (A) Within each four-point group, no two points were simultaneously within 20 pixels of each other in both the horizontal and vertical directions. (B) All points were located more than 20 pixels away from image borders to avoid boundary-related artifacts. (C) To prevent ambiguity near object edges, each point was placed at least 5 pixels away from segmentation boundaries. For the KITTI images, these boundaries were estimated using the Segment Anything Model (SAM)~\cite{kirillov2023segany}, whereas for the NYU images, we used the segmentation maps provided with the dataset.

Candidate points were selected separately for each four-point group through an iterative sampling procedure. A randomly selected point $p_i$ was compared with all previously chosen points $[p_0, \dots, p_{i-1}]$ to verify that it satisfied the three spatial constraints. If any condition was violated, nearby pixels within the same segmentation region were explored as alternative candidates. This local search strategy reduced the oversampling of large homogeneous areas such as roads or walls. When no valid candidate could be found within a given segment, the selection process was reinitialized until a complete set of sixteen valid points was obtained. The final point sets were subsequently used to collect absolute depth judgments from human participants. 

\subsection{Human data collection procedure}
\label{app:human-data}
Participants were adults aged 20--49 years recruited through an online crowdsourcing platform. We recruited 1,847 participants for KITTI and 1,782 for NYU. Before starting the experiment, participants were instructed to view the stimuli binocularly, to disable dark or night mode on their browsers, and to view the stimuli in full-screen mode. To standardize the physical size of stimuli across different display devices, participants adjusted an on-screen rectangle to match the size of a credit card held in their hand. This calibration ensured consistent stimulus scaling based on individual screen pixel density, resulting in an image size of 18.6 cm × 5.63 cm (KITTI) and 9.6 cm $\times$ 7.2 cm (NYU). Participants were also instructed to maintain a consistent viewing distance of 60 cm, which resulted in a stimulus visual angle of approximately 17.6$^\circ$ $\times$ 5.37$^\circ$ (KITTI) and 9.15$^\circ$ $\times$ 6.87$^\circ$ (NYU). Although our online setup did not allow for detailed control of viewing conditions or individual visual acuity, averaging responses across a large sample reduced potential variability from such uncontrolled factors.

Prior to the main trials, participants completed five practice trials with feedback of ground truth values to ensure task comprehension. During the experiment, they were presented with stimuli containing four labeled target points (A--D) and were asked to estimate the distance of each point from their viewpoint, imagining themselves positioned at the camera location. For reflective or transparent surfaces (e.g., mirrors, glass), participants reported the perceived distance to the physical surface itself. Each participant completed 28 or 32 randomized trials, and every stimulus set was evaluated by at least 20 individuals before participant screening. 

To maintain data quality, we excluded participants whose responses showed low reliability. For each participant, we computed the Pearson correlation between the participant's responses and the group-median responses across the same evaluation points. The group median was calculated using all participants assigned to the stimulus set. Reliability scores were then pooled across participants, and the lower outlier threshold was defined as $Q_1 - 1.5\times \mathrm{IQR}$, where $Q_1$ denotes the first quartile and IQR denotes the interquartile range of the pooled reliability-score distribution. Participants whose reliability scores fell below this threshold were excluded from subsequent analyses. After screening, 1,547 KITTI participants and 1,483 NYU participants remained. The number of valid responses per image averaged 18.87 for KITTI and 18.09 for NYU, with ranges of 14--27 and 12--26, respectively.

\subsection{Target DNN models}
\label{app:target-model}
Table~\ref{tab:model-list} summarizes the 69 monocular depth estimators adopted in this study. The collection spans diverse strategies (supervised, self-supervised, hybrid, and generative), backbone architectures (CNNs, Transformers, CNN--Transformer hybrids, and diffusion-based models), datasets, and parameter sizes. Supervised approaches employ ground-truth depth supervision, while self-supervised models learn from photometric consistency across stereo or monocular sequences. Hybrid methods integrate both paradigms: `hybrid (disparity)' models, such as DistDepth~\cite{wu2022toward} and SCDepth-v3~\cite{sun2023sc}, combine teacher-student supervision with disparity-based self-training; `hybrid (semantic)' models, including Depth-Anything~\cite{yang2024depth} and DINOv2~\cite{oquab2023dinov2}, leverage semantic features from large-scale contrastive pretraining. Generative models, such as EVP~\cite{lavreniuk2023evp} and Marigold~\cite{ke2023repurposing}, incorporate diffusion-based representations via Stable Diffusion~\cite{rombach2022high} coupled with supervised decoders for depth estimations.

Our evaluation covers DNNs trained not only on NYU~\cite{silberman2012indoor} and KITTI~\cite{uhrig2017sparsity} but also on broader datasets to examine generalization across domains. These include indoor datasets such as Bonn~\cite{palazzolo2019refusion} and TUM~\cite{sturm2012benchmark}, as well as outdoor datasets like DDAD~\cite{guizilini20203d}. Mixed-domain models such as MiDaS~\cite{ranftl2021vision,ranftl2020towards} and ZoeDepth~\cite{bhat2023zoedepth} were pretrained on large composite datasets (e.g., MIX6, MIX10) combining synthetic and real imagery from diverse sources. Depth-Anything~\cite{yang2024depth} and Metric3D~\cite{yin2023metric3d} further extend this paradigm by integrating numerous labeled and pseudo-labeled datasets across both indoor and outdoor environments.

All pretrained weights were obtained from publicly available repositories. Minimal code adjustments were made solely to ensure compatibility with our hardware (GALLERIA UL9C-R49-6, RTX 4090 GPU) without altering model architectures or pretrained parameters. Model suffixes indicate dataset and architectural variations: `-K', `-N', `-B', `-D', and `-T' correspond to KITTI, NYU, Bonn, DDAD, and TUM, respectively; `-NK' denotes dual fine-tuning on KITTI and NYU. Backbone abbreviations include `cn' (CNN), `tf' (Transformer), `hy' (hybrid), and `df' (diffusion). Parameter counts are expressed as, for example, `-49M' or `-87M'.

For evaluation consistency, DNN outputs were categorized as absolute depth, relative depth ($^\dagger$), or disparity ($^\ddagger$), stored in 16-bit (uint16) format. When we calculated scale-recovered data from DNN outputs, we performed per-image linear regression between predictions ($z_\mathrm{out}$) and ground truth ($z_\mathrm{GT}$) for sixteen evaluation points, fitting $z_\mathrm{GT} = s*z_\mathrm{out} + t$ to estimate optimal scaling $(s^*)$ and shifting $(t^*)$ factors. Transformed outputs were computed as $z'_\mathrm{out} = s^**z_\mathrm{out} + t^*$. For disparity-based models, predicted disparity was aligned to
inverse of ground truth using the same scale-and-shift fitting procedure. The aligned disparity was then inverted to obtain pseudo-absolute depth. The resulting depth estimates were clipped to the dataset-specific response ranges used in the human experiment (0--10 m for NYU Depth V2 and 0--100 m for KITTI) ensuring that human and model estimates were compared within the same range. We therefore applied this alignment procedure consistently to all human and model outputs used in the scale-recovered analyses.

\subsection{Computing environment}
Depth inference with the 69 MDE models was performed on a GALLERIA UL9C-R49-6 computer running Windows 11 Pro 25H2, equipped with a 13th Gen Intel Core i9-13900HX CPU, 64 GB of RAM, and an NVIDIA GeForce RTX 4090 GPU with 24 GB of GPU memory. All subsequent data processing, statistical analyses, and figure generation were performed on a MacBook Pro running macOS 15.5, equipped with an Apple M2 Pro processor and 32 GB of RAM. Detailed software dependencies and package versions are provided in the corresponding environment specification files in each subdirectory.

\begin{table}[htbp]
    \caption{Target 69 DNN model list. The model outputs were categorized based on the type of depth information provided: absolute depth (no symbol), relative depth (indicated by $^\dagger$), and disparity (indicated by $^\ddagger$). Full version of this table is available as the \texttt{model\_list.csv} file.}
    \centering
    \footnotesize
    \label{tab:model-list}
    \begin{tabular}{p{5cm}ccccp{1.5cm}c}
    \hline
    \textbf{Models} & \textbf{Strategy} & \textbf{Backbone} & \multicolumn{3}{c}{\textbf{Dataset}}  & \textbf{Parameters} \\ 
    & & & KITTI & NYU & Others &    \\ \hline
    AdaBins-cn-K~\cite{bhat2021adabins} & supervised & CNN & $\checkmark$ & - & - & 78,257,238 \\
    AdaBins-cn-N~\cite{bhat2021adabins} & supervised & CNN & - & $\checkmark$ & - & 78,257,238 \\
    BinsFormer-tf-49M-N~\cite{li2022binsformer} & supervised & Transformer & - & $\checkmark$ & - & 49,231,191 \\
    BinsFormer-tf-255M-N~\cite{li2022binsformer} & supervised & Transformer & - & $\checkmark$ & - & 254,625,329 \\
    BTS-cn-K~\cite{lee2019big} & supervised & CNN & $\checkmark$  & - & - & 47,481,969 \\
    BTS-cn-N~\cite{lee2019big} & supervised & CNN & - & $\checkmark$ & - & 47,481,969 \\
    Depth-Anything-tf-25M~\cite{yang2024depth}$^\dagger$ & hybrid (semantic) & Transformer & - &  -  & 14 datasets & 24,785,089 \\
    Depth-Anything-tf-97M~\cite{yang2024depth}$^\dagger$ & hybrid (semantic) & Transformer & - &  -  & 14 datasets & 97,470,785 \\
    Depth-Anything-tf-335M~\cite{yang2024depth}$^\dagger$ & hybrid (semantic) & Transformer & - &  -  & 14 datasets & 335,315,649 \\
    DepthFormer-hy-48M-N~\cite{li2022depthformer} & supervised & hybrid & - & $\checkmark$ & - & 47,612,029 \\
    DepthFormer-hy-274M-K~\cite{li2022depthformer} & supervised & hybrid & $\checkmark$ & - & - & 273,732,631 \\
    DepthFormer-hy-274M-N~\cite{li2022depthformer} & supervised & hybrid & - & $\checkmark$ & - & 273,732,631 \\
    DINOv2-tf-36M-K~\cite{oquab2023dinov2} & hybrid (semantic) & Transformer & $\checkmark$ & - & 1.2B data & 36,017,953 \\
    DINOv2-tf-36M-N~\cite{oquab2023dinov2} & hybrid (semantic) & Transformer & - & $\checkmark$ & 1.2B data & 36,017,953 \\
    DINOv2-tf-111M-K~\cite{oquab2023dinov2} & hybrid (semantic) & Transformer & $\checkmark$ & - & 1.2B data & 110,774,401 \\
    DINOv2-tf-111M-N~\cite{oquab2023dinov2} & hybrid (semantic) & Transformer & - & $\checkmark$ & 1.2B data & 110,774,401 \\
    DINOv2-tf-339M-K~\cite{oquab2023dinov2} & hybrid (semantic) & Transformer & $\checkmark$ & - & 1.2B data & 338,558,657 \\
    DINOv2-tf-339M-N~\cite{oquab2023dinov2} & hybrid (semantic) & Transformer & - & $\checkmark$ & 1.2B data & 338,558,657 \\
    DINOv2-tf-1198M-K~\cite{oquab2023dinov2} & hybrid (semantic) & Transformer & $\checkmark$ & - & 1.2B data & 1,198,281,537 \\
    DINOv2-tf-1198M-N~\cite{oquab2023dinov2} & hybrid (semantic) & Transformer & - & $\checkmark$ & 1.2B data & 1,198,281,537 \\
    DistDepth-cn~\cite{wu2022toward} & hybrid (disparity) & CNN & - & - & 3 datasets & 69,206,908 \\
    Eigen2014-cn~\cite{eigen2014depth}$^\dagger$ & supervised & CNN & - & $\checkmark$ & - & 240,833,218 \\
    EVP-df~\cite{lavreniuk2023evp} & generative & Diffusion & - & $\checkmark$ & LAION-5B & 933,814,544 \\
    GasMono-tf~\cite{zhao2023gasmono} & self-supervised & Transformer & - & $\checkmark$ & - & 28,000,697 \\
    GLPDepth-hy-K~\cite{kim2022global} & supervised & hybrid & $\checkmark$ & - & - & 61,220,903 \\
    GLPDepth-hy-N~\cite{kim2022global} & supervised & hybrid & - & $\checkmark$ & - & 61,220,903 \\
    IndoorDepth-cn~\cite{fan2023deeper}$^\dagger$ & self-supervised & CNN & - & $\checkmark$ & - & 14,840,507 \\
    Marigold-df~\cite{ke2023repurposing}$^\dagger$ & generative & Diffusion & - & - & Hypersim, Virtual KITTI & 1,289,963,947 \\
    Metric3D-cn~\cite{yin2023metric3d} & supervised & CNN & - & - & 11 datasets & 203,243,090 \\
    MiDaS-tf-BEiT-B384-112M~\cite{birkl2023midas}$^\ddagger$ & supervised & Transformer & - & - & 10 datasets & 111,531,689 \\
    MiDaS-tf-BEiT-L384-344M~\cite{birkl2023midas}$^\ddagger$ & supervised & Transformer & - & - & 10 datasets & 344,338,601 \\
    MiDaS-tf-BEiT-L512-345M~\cite{birkl2023midas}$^\ddagger$ & supervised & Transformer & - & - & 10 datasets & 345,014,441 \\
    MiDaS-tf-LeViT-224-51M~\cite{birkl2023midas}$^\ddagger$ & supervised & Transformer & - & - & 10 datasets & 50,630,837 \\
    MiDaS-tf-NextViT-L384-72M~\cite{birkl2023midas}$^\ddagger$ & supervised & Transformer & - & - & 10 datasets & 72,260,713 \\ \hline
    \end{tabular}
\end{table}

\begin{table}[htbp]
    \ContinuedFloat
    \caption{Target 69 DNN model list (continued).}
    \label{tab:model-list-continued}
    \centering
    \begin{tabular}{p{5cm}ccccp{1.5cm}c}
    \hline
    \footnotesize
    \textbf{Models} & \textbf{Strategy} & \textbf{Backbone}
    & \multicolumn{3}{c}{\textbf{Dataset}} & \textbf{Parameters} \\
    & & & KITTI & NYU & Others & \\ \hline
    MiDaS-tf-Swin-L384-213M~\cite{birkl2023midas}$^\ddagger$ & supervised & Transformer & - & - & 10 datasets & 213,407,453 \\
    MiDaS-tf-Swin2-B384-102M~\cite{birkl2023midas}$^\ddagger$ & supervised & Transformer & - & - & 10 datasets & 102,378,913 \\
    MiDaS-tf-Swin2-L384-213M~\cite{birkl2023midas}$^\ddagger$ & supervised & Transformer & - & - & 10 datasets & 213,411,869 \\
    MiDaS-tf-Swin2-T256-42M~\cite{birkl2023midas}$^\ddagger$ & supervised & Transformer & - & - & 10 datasets & 41,701,331 \\
    MiDaS-hy-H384-123M~\cite{ranftl2021vision}$^\ddagger$ & supervised & hybrid & - & - & 10 datasets & 123,146,985 \\
    MiDaS-hy-H384-123M-K~\cite{ranftl2021vision}$^\dagger$ & supervised & hybrid & $\checkmark$ & - & 10 datasets & 123,146,985 \\
    MiDaS-hy-H384-123M-N~\cite{ranftl2021vision} & supervised & hybrid & - & $\checkmark$ & 10 datasets & 123,146,985 \\ 
    MiDaS-cn-L384-344M~\cite{ranftl2021vision}$^\ddagger$ & supervised & CNN & - & - & 10 datasets & 344,055,465 \\
    MiDaSv21-cn-L384-105M~\cite{ranftl2020towards}$^\ddagger$ & supervised & CNN & - & - & 6 datasets & 105,362,945 \\
    MiDaSv21-cn-S256-21M~\cite{ranftl2020towards}$^\ddagger$ & supervised & CNN & - & - & 6 datasets & 21,320,545 \\
    MIM-tf-87M-K~\cite{xie2023revealing} & supervised & Transformer & $\checkmark$ & - & - & 87,219,257 \\
    MIM-tf-87M-N~\cite{xie2023revealing} & supervised & Transformer & - & $\checkmark$ & - & 87,219,257 \\
    NeWCRFs-tf-K~\cite{yuan2022new} & supervised & Transformer & $\checkmark$ & - & - & 270,444,877 \\
    NeWCRFs-tf-N~\cite{yuan2022new} & supervised & Transformer & - & $\checkmark$ & - & 270,444,877 \\
    Omnidata-tf~\cite{eftekhar2021omnidata}$^\dagger$ & supervised & Transformer & - & - & 7 datasets & 123,146,985 \\
    P2Net-cn-3frames~\cite{yu2020p}$^\ddagger$ & self-supervised & CNN & - & $\checkmark$ & - & 14,840,217 \\
    P2Net-cn-5frames~\cite{yu2020p}$^\ddagger$ & self-supervised & CNN & - & $\checkmark$ & - & 14,840,217 \\
    PackNet-cn~\cite{guizilini20203d}$^\dagger$ & self-supervised & CNN & $\checkmark$ & - & - & 128,294,020 \\
    SCDepth-v1-cn-D~\cite{bian2021unsupervised}$^\dagger$ & self-supervised & CNN & - & - & DDAD & 14,842,236 \\
    SCDepth-v1-cn-K~\cite{bian2021unsupervised}$^\dagger$ & self-supervised & CNN & $\checkmark$ & - & - & 14,842,236 \\
    SCDepth-v2-cn-N~\cite{bian2021auto}$^\dagger$ & self-supervised & CNN & - & $\checkmark$ & - & 14,842,236 \\
    SCDepth-v3-cn-B~\cite{sun2023sc}$^\dagger$ & hybrid (disparity) & CNN & - & - & Bonn & 14,842,236 \\
    SCDepth-v3-cn-D~\cite{sun2023sc}$^\dagger$ & hybrid (disparity) & CNN & - & - & DDAD & 14,842,236 \\
    SCDepth-v3-cn-K~\cite{sun2023sc}$^\dagger$ & hybrid (disparity) & CNN & $\checkmark$ & - & - & 14,842,236 \\
    SCDepth-v3-cn-N~\cite{sun2023sc}$^\dagger$ & hybrid (disparity) & CNN & - & $\checkmark$ & - & 14,842,236 \\
    SCDepth-v3-cn-T~\cite{sun2023sc}$^\dagger$ & hybrid (disparity) & CNN & - & - & TUM & 14,842,236 \\
    StructDepth-cn~\cite{li2021structdepth}$^\ddagger$ & self-supervised & CNN & - & $\checkmark$ & - & 14,840,217 \\
    TrainFlow-cn-K~\cite{zhao2020towards} & self-supervised & CNN & $\checkmark$ & - & - & 19,974,541 \\
    TrainFlow-cn-N~\cite{zhao2020towards} & self-supervised & CNN & - & $\checkmark$ & - & 19,974,541 \\
    TransDepth-hy-K~\cite{yang2021transformer} & supervised & hybrid & $\checkmark$ & - & - & 247,399,587 \\
    TransDepth-hy-N~\cite{yang2021transformer} & supervised & hybrid & - & $\checkmark$ & - & 247,399,587 \\
    ZeroDepth-tf~\cite{guizilini2023towards} & supervised & Transformer & - & - & 5 datasets & 232,591,380 \\
    ZoeDepth-tf-K~\cite{bhat2023zoedepth} & supervised & Transformer & $\checkmark$ & - & 10 datasets & 344,820,487 \\
    ZoeDepth-tf-N~\cite{bhat2023zoedepth} & supervised & Transformer & - & $\checkmark$ & 10 datasets & 334,816,746 \\
    ZoeDepth-tf-NK~\cite{bhat2023zoedepth} & supervised & Transformer & $\checkmark$ & $\checkmark$ & 10 datasets & 346,100,355 \\ \hline
    \end{tabular}
\end{table}

\clearpage

\section{Supplemental Results}
\subsection{Similarity matrices among humans and DNNs}
\label{app:mds}
\begin{figure}[htb]
  \centering
  \includegraphics[keepaspectratio,width=1.00\linewidth]{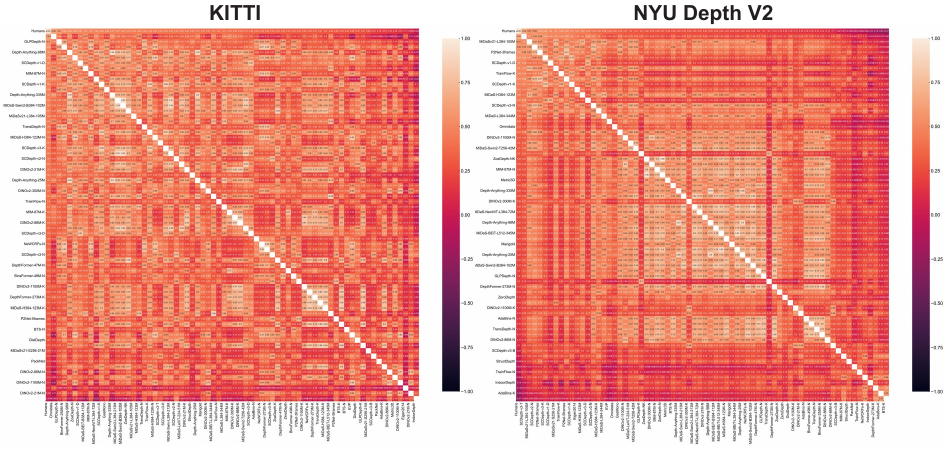}
  \caption{Heatmaps of pairwise error-pattern similarity among humans and MDE models for KITTI (left) and NYU (right). Each cell represents the Pearson partial correlation between a pair of depth-estimation error patterns after controlling for physical ground-truth depth.}
  \label{fig:similarity-models}
\end{figure}

Figure~\ref{fig:similarity-models} compares the pairwise error-pattern similarity matrices. Both matrices revealed clear block structures rather than uniform similarity across models. These blocks included mutually similar high-accuracy models as well as other clusters that appeared to correspond to shared model families and  training datasets. This organization is consistent with systematic residual error patterns being shared within particular groups of models.

\subsection{Additional cross-dataset analyses}
\label{app:cross-data}
\begin{figure*}[htb]
  \centering
  \includegraphics[keepaspectratio,width=1.00\linewidth]{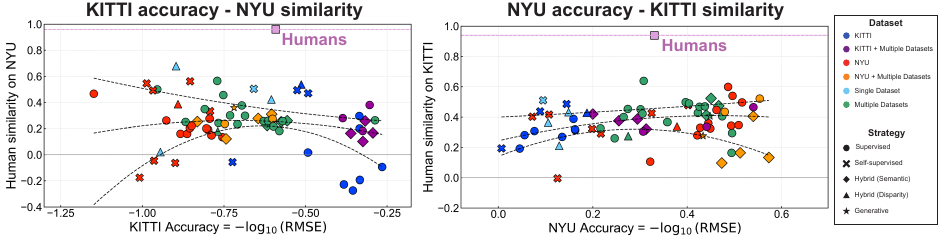}
  \caption{Cross-dataset relationships between metric accuracy and human--model similarity. The left panel shows KITTI accuracy versus human similarity on NYU, and the right panel shows NYU accuracy versus human similarity on KITTI. The three curves show quadratic quantile-regression fits at the 0.20, 0.50, and 0.80 quantiles across models. The human benchmark points were not included in the regression fits.}
  \label{fig:similarity-cross-accuracy}
\end{figure*}

Figure~\ref{fig:similarity-cross-accuracy} examines whether metric accuracy in one domain predicts human--model similarity in the other. As shown in the left panel, higher KITTI accuracy was associated with a weak downward trend in human similarity on NYU, whereas the right panel showed a weak upward trend between NYU accuracy and human similarity on KITTI. However, in both cross-dataset analyses, the models remained widely dispersed around the fitted conditional median, and the central 20th--80th percentile range did not narrow toward the highest-accuracy end. For KITTI accuracy predicting NYU similarity, the interval width increased from its minimum of $0.12$ at an intermediate log-accuracy of $-0.57$ to $0.35$ at the highest observed log-accuracy of $-0.27$. Similarly, for NYU accuracy predicting KITTI similarity, the width increased from $0.12$ at an intermediate log-accuracy of $0.25$ to $0.36$ at the highest observed log-accuracy of $0.57$. Thus, although the fitted median trends differed in direction across the two cross-domain comparisons, higher accuracy in one dataset did not consistently constrain or improve human similarity in the other,  indicating that human-likeness depends on the interaction between model-specific biases and domain-specific scene statistics.


\subsection{Comparing individual datasets and backbone variations}
\label{app:dnn-individual}
\begin{figure*}[htb]
\begin{center}
\includegraphics[keepaspectratio,width=1.00\linewidth]{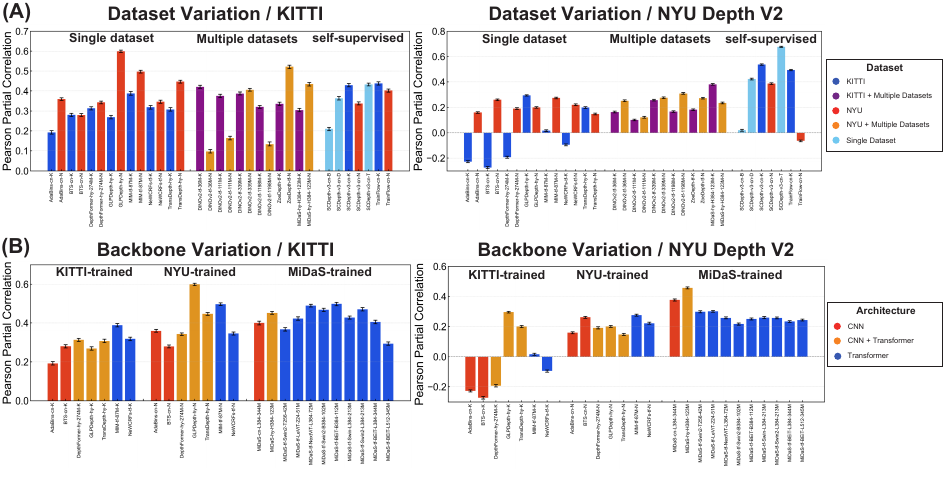}
\end{center}
\caption{Human--model similarity evaluated on KITTI (left) and NYU (right). (A) Training-dataset variants, with colors indicating training-dataset categories. (B) Backbone-architecture variants, with colors indicating architecture types. Error bars indicate 95\% confidence intervals obtained from bootstrap resamples of human responses.}
\label{fig:dnn-individuals}
\end{figure*}

To explore factors associated with variation in human--model similarity beyond overall accuracy, we conducted comparisons within selected model families. We first compared identical or closely related models trained or fine-tuned on different datasets (Figure~\ref{fig:dnn-individuals}A). Among model pairs trained directly on KITTI or NYU, NYU-trained variants often showed higher human similarity than their KITTI-trained counterparts in both evaluations. This contrast was particularly pronounced in NYU, where several KITTI-trained variants exhibited near-zero or negative similarity, whereas the corresponding NYU-trained variants showed positive correlations. However, the pattern was not universal: reversals occurred in several model families, and the direction of the difference also varied among models pretrained on multiple datasets and subsequently fine-tuned on KITTI or NYU. Thus, training data clearly influenced human similarity, but the results could not be explained by simple matching between the training and evaluation domains.

Next, we assessed the influence of learning strategies by analyzing models with different forms of supervision, including the SCDepth-v3 family~\cite{sun2023sc}, which employs disparity-based self-supervision (Figure~\ref{fig:dnn-individuals}A). Some variants within the SCDepth-v3 family achieved relatively high human similarity, whereas other variants using similar learning strategies showed substantially lower values. Similarly, generative and hybrid-semantic models did not consistently outperform other strategies. This variability underscores that self-supervision or any other training strategy alone does not guarantee human-like depth-estimation behavior.

We next compared backbone architectures within benchmark-trained model groups and the MiDaS family (Figure~\ref{fig:dnn-individuals}B). Architecture changes were associated with moderate differences in human similarity, but no consistent ordering emerged across datasets or model families. Some Transformer-based models achieved high similarity in the KITTI evaluation, whereas CNN--Transformer hybrids were comparable or superior in several other comparisons. In the NYU evaluation, for example, the MiDaS hybrid model showed greater human similarity than the corresponding Transformer variants. Overall, these descriptive comparisons suggest that human-like error patterns depend jointly on training data, fine-tuning domain, architecture, and model family, rather than being reliably predicted by any single model characteristic.

\subsection{Additional qualitative comparisons of human and model depth-estimation biases}
\label{app:human-examples}
To obtain a representative set of images spanning the observed range of human--model similarity, we selected images according to their image-level similarity between human judgments and model estimates. For each image and model, we first computed the Pearson partial correlation between the model predictions and the mean human judgments across the 16 evaluation locations while controlling for ground-truth depth. These model-wise, image-level correlations were Fisher $z$-transformed and averaged across models to obtain a  similarity score for each image. Images were then divided into 10 approximately equally sized bins ordered from lower to higher similarity scores, and the image closest to the median score within each bin was selected. Two of the resulting 10 examples are shown in the main manuscript, and the remaining eight are presented in this Appendix. The same images were used for the absolute-output comparisons, allowing direct comparison across output representations.

\begin{figure*}[htb]
  \centering
  \includegraphics[keepaspectratio,width=0.95\linewidth]{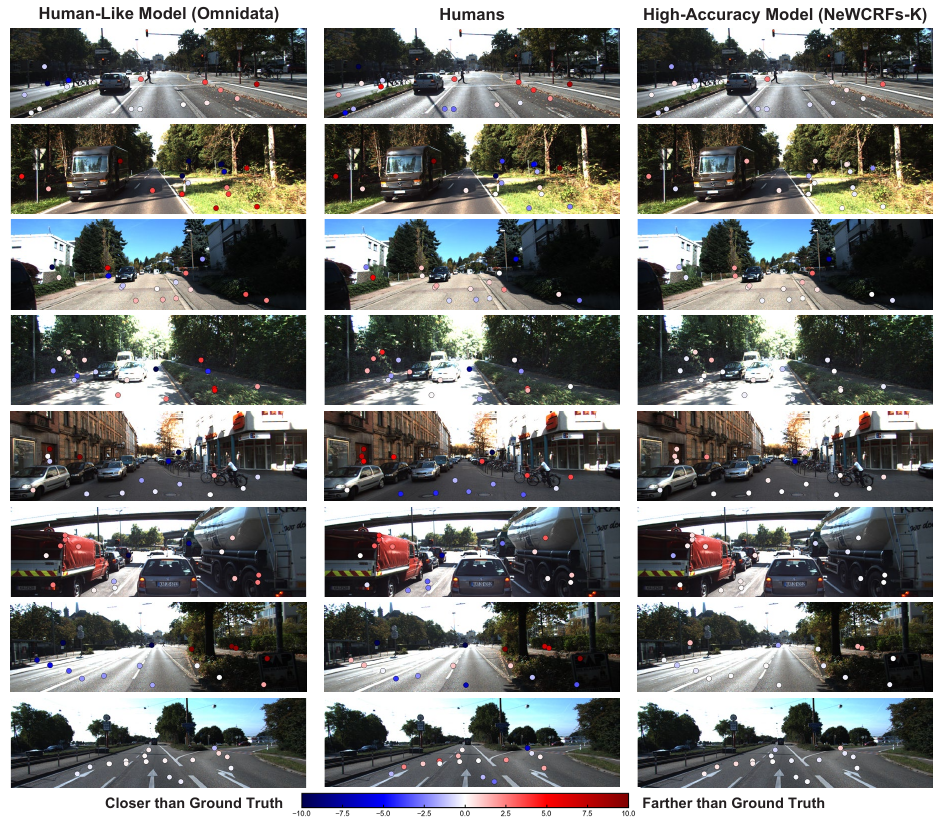}
  \caption{Additional qualitative comparisons of human and MDE model depth-estimation biases for KITTI using scale-recovered estimates. The columns show a human-like model (left), humans (center), and the high-accuracy model (right). Marker colors represent the signed depth error in meters; blue and red indicate smaller and larger estimated depths, respectively.}
  \label{fig:examples-kitti-add}
\end{figure*}

\begin{figure*}[htb]
  \centering
  \includegraphics[keepaspectratio,width=0.95\linewidth]{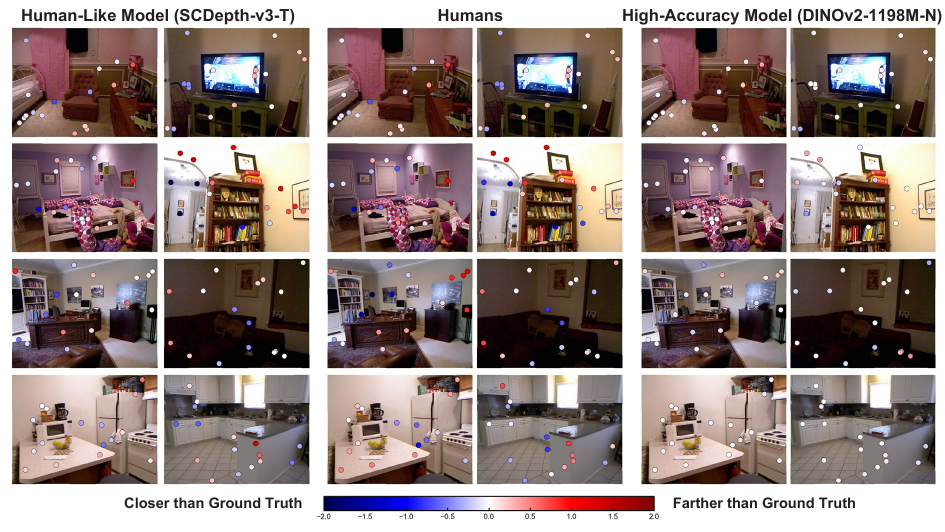}
  \caption{Additional qualitative comparisons of human and MDE model depth-estimation biases for NYU using scale-recovered estimates. The columns show a human-like model (left), humans (center), and the high-accuracy model (right). Marker colors represent the signed depth error in meters; blue and red indicate smaller and larger estimated depths, respectively.}
  \label{fig:examples-nyu-add}
\end{figure*}

\begin{figure*}[htb]
  \centering
  \includegraphics[keepaspectratio,width=0.95\linewidth]{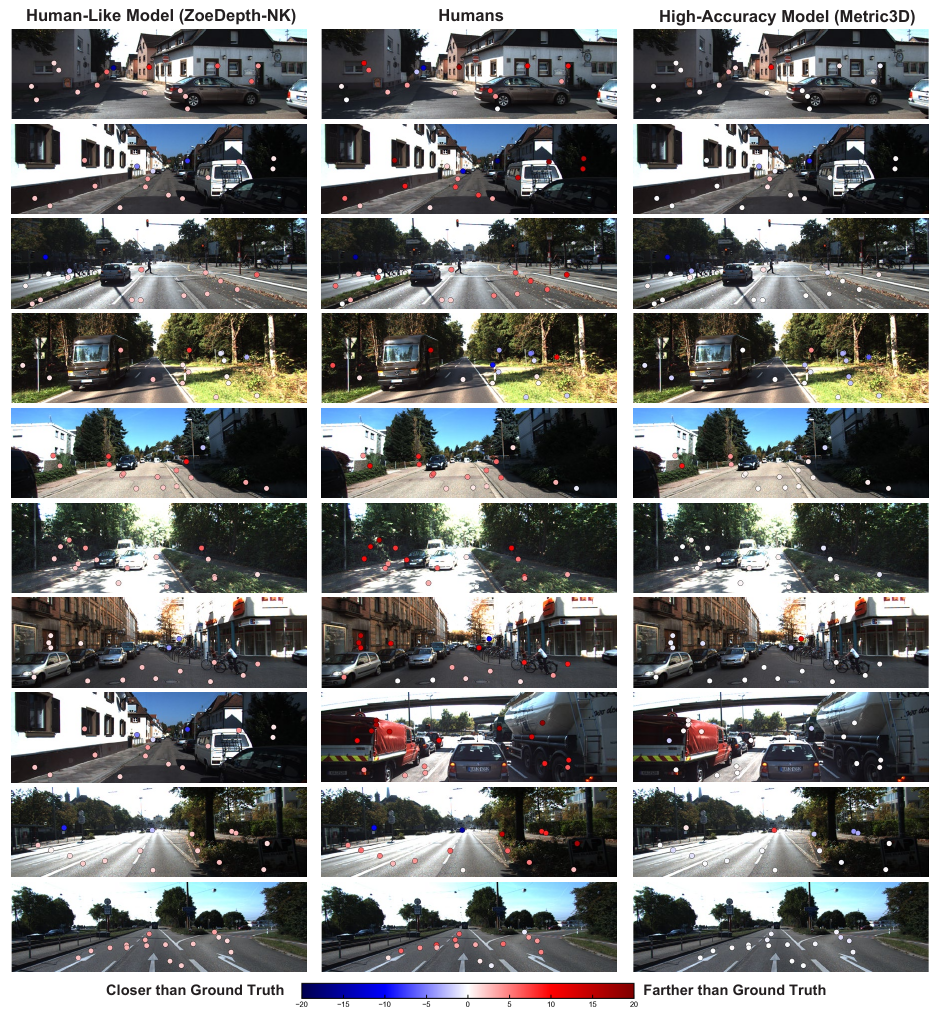}
  \caption{Qualitative comparisons of human and MDE model depth-estimation biases for KITTI using absolute estimates. The columns show a human-like model (left), humans (center), and the high-accuracy model (right). Marker colors represent the signed depth error in meters; blue and red indicate smaller and larger estimated depths, respectively.}
  \label{fig:examples-kitti-abs}
\end{figure*}

\begin{figure*}[htb]
  \centering
  \includegraphics[keepaspectratio,width=0.95\linewidth]{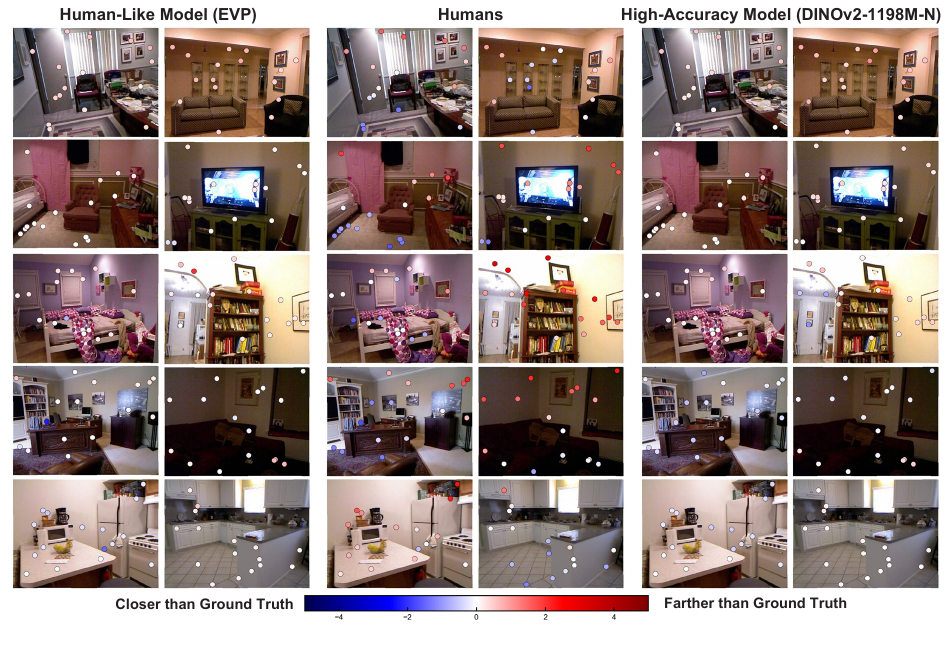}
  \caption{Qualitative comparisons of human and MDE model depth-estimation biases for NYU using absolute estimates. The columns show a human-like model (left), humans (center), and the high-accuracy model (right). Marker colors represent the signed depth error in meters; blue and red indicate smaller and larger estimated depths, respectively.}
  \label{fig:examples-nyu-abs}
\end{figure*}

Figures~\ref{fig:examples-kitti-add} and \ref{fig:examples-nyu-add} present additional qualitative comparisons based on scale-recovered estimates, whereas Figures~\ref{fig:examples-kitti-abs} and \ref{fig:examples-nyu-abs} show the corresponding absolute-depth outputs. For each dataset, we selected the models with the highest human similarity and the models with the highest metric accuracy, and visualized the same two models across all selected images. For the absolute-output comparisons, candidate models were restricted to those producing absolute metric-depth estimates. We additionally excluded models whose native metric scale was incompatible with the evaluation domain, such as KITTI-specific models exhibiting a clear scale mismatch when evaluated on NYU. From the remaining models, we selected the model with the highest human similarity and the model with the highest metric accuracy for each dataset.

Across the KITTI examples, humans and the selected human-like models showed partially corresponding deviations on roads, distant scene regions, vehicles, and other upright structures, whereas the high-accuracy models often exhibited differently organized residual patterns. In the NYU examples, humans and human-like models sometimes shared overestimation on walls, doors, shelves, and furniture. However, the lower-region bias was generally more apparent and consistent in the human estimates than in either model. The absolute-output visualizations additionally retained global non-linear biases, including compression of distant depth. Importantly, after global scale and shift were removed, human-like models still partially reproduced aspects of the point-wise sign and spatial organization of human residuals. These examples illustrate residual correspondences that remain after global scale and shift have been removed, indicating that the selected models partially reproduce a subset of the biases observed in human depth judgments.



\end{document}